%% file: main.tex
\documentclass[letterpaper, 10 pt, conference]{ieeeconf}

\IEEEoverridecommandlockouts                              

\usepackage{microtype}
\usepackage{amsmath,amssymb,amsfonts}
\usepackage{algorithmic}
\usepackage{graphicx}
\usepackage{textcomp}
\usepackage{xcolor}
\usepackage{subfig}
\usepackage{gensymb}
\usepackage{caption}
\usepackage{hyperref}
\usepackage{mwe}
\usepackage{amsfonts}
\usepackage{amssymb}
\usepackage{verbatim}
\usepackage{amsbsy}
\usepackage{siunitx}
\usepackage{tabularx, booktabs}
\usepackage{dblfloatfix}
\usepackage{cite}
\usepackage{cleveref}

\usepackage{algorithm, algorithmic}
\usepackage[autolanguage]{numprint}

\usepackage{spreadtab}
\usepackage{multirow}
\usepackage{cite}

\usepackage{hyperref}
\usepackage{xcolor}
\usepackage{algorithmic}
\usepackage{textcomp}
\usepackage{xcolor}
\usepackage{subfig}
\usepackage{gensymb}
\usepackage{caption}

\usepackage{mwe}
\usepackage{amsfonts}
\usepackage{amssymb}
\usepackage{verbatim}
\usepackage{amsbsy}
\usepackage{siunitx}
\usepackage{tabularx, booktabs}
\usepackage{soul}
\usepackage{tikz}
\usetikzlibrary{calc}

\makeatletter
\newif\if@anonymize

\@anonymizetrue    

\if@anonymize
  \newcommand{\highlight@DoHighlight}{
    \fill [outer sep = -15pt, inner sep = 0pt, color=black]
          ($(begin highlight)+(0,8pt)$) rectangle ($(end highlight)+(0,-3pt)$) ;
  }

  \newcommand{\highlight@BeginHighlight}{
    \coordinate (begin highlight) at (0,0) ;
  }

  \newcommand{\highlight@EndHighlight}{
    \coordinate (end highlight) at (0,0) ;
  }

  \newdimen\highlight@previous
  \newdimen\highlight@current
  \newlength{\item@width}

  \DeclareRobustCommand*\anonymize{%
    \SOUL@setup
    \def\SOUL@preamble{%
      \begin{tikzpicture}[overlay, remember picture]
        \highlight@BeginHighlight
        \highlight@EndHighlight
      \end{tikzpicture}%
    }%
    \def\SOUL@postamble{%
      \begin{tikzpicture}[overlay, remember picture]
        \highlight@EndHighlight
        \highlight@DoHighlight
      \end{tikzpicture}%
    }%
    \def\SOUL@everyhyphen{%
      \discretionary{%
        \SOUL@setkern\SOUL@hyphkern
        \SOUL@sethyphenchar
        \tikz[overlay, remember picture] \highlight@EndHighlight ;%
      }{%
      }{%
        \SOUL@setkern\SOUL@charkern
      }%
    }%
    \def\SOUL@everyexhyphen##1{%
      \SOUL@setkern\SOUL@hyphkern
      \settowidth{\item@width}{##1}%
      \makebox[\item@width]{}%
      \discretionary{%
        \tikz[overlay, remember picture] \highlight@EndHighlight ;%
      }{%
      }{%
        \SOUL@setkern\SOUL@charkern
      }%
    }%
    \def\SOUL@everysyllable{%
      \begin{tikzpicture}[overlay, remember picture]
        \path let \p0 = (begin highlight), \p1 = (0,0) in \pgfextra
          \global\highlight@previous=\y0
          \global\highlight@current =\y1
        \endpgfextra (0,0) ;
        \ifdim\highlight@current < \highlight@previous
          \highlight@DoHighlight
          \highlight@BeginHighlight
        \fi
      \end{tikzpicture}%
      \settowidth{\item@width}{\the\SOUL@syllable}%
      \makebox[\item@width]{}%
      \tikz[overlay, remember picture] \highlight@EndHighlight ;%
    }%
    \SOUL@
  }
\else
  \newcommand{\anonymize}[1]{#1}
\fi
\makeatother 

\title{\LARGE \bf
Deterministic and Stochastic Analysis of Deep Reinforcement Learning for Low Dimensional Sensing-based Navigation of Mobile Robots}

\author{Ricardo B. Grando$^{1,2}$, Junior C. de Jesus$^{2}$, Victor A. Kich$^{3}$, Alisson H. Kolling$^{3}$, \\Rodrigo S. Guerra$^{3}$, Paulo L. J. Drews-Jr$^{2}$
\thanks{$^{1}$Ricardo B. Grando is with the Technological University of Uruguay. E-mail: {\tt\small ricardo.bedin@utec.edu.uy}}
\thanks{$^{2}$Ricardo B. Grando, Junior C. de Jesus and P. L. J. Drews-Jr are with NAUTEC - Intelligent Robotics and Automation Group, Center for Computational Science,  Federal University of Rio Grande - FURG, RS, Brazil. E-mail: {\tt\small paulodrews@furg.br}}
\thanks{$^{3}$Victor A. Kich, Alisson H. Kolling and Rodrigo S. Guerra are with the Universidade Federal de Santa Maria - UFSM, RS, Brazil. E-mail: {\tt\small rodrigo.guerra@ufsm.br}}
}

\begin{document}

\maketitle
\thispagestyle{empty}
\pagestyle{empty}

\input{sections/0_0_abstract.tex}
\vspace{-3mm}
\input{sections/0_1_supplementary_material.tex}
\vspace{-2.5mm}
\input{sections/1_introduction.tex}
\input{sections/2_related_works.tex}
\input{sections/3_methodology.tex}

\input{sections/4_experimental_results.tex}
\input{sections/5_discussion.tex}
\input{sections/6_conclusion.tex}
\input{sections/7_acknowledgment.tex}
\input{sections/8_references.tex}

\end{document}

%% file: sections/0_0_abstract.tex
\begin{abstract}
Deterministic and Stochastic techniques in Deep Reinforcement Learning (Deep-RL) have become a promising solution to improve motion control and the decision-making tasks for a wide variety of robots. Previous works showed that these Deep-RL algorithms can be applied to perform mapless navigation of mobile robots in general. However, they tend to use simple sensing strategies since it has been shown that they perform poorly with a high dimensional state spaces, such as the ones yielded from image-based sensing. This paper presents a comparative analysis of two Deep-RL techniques - Deep Deterministic Policy Gradients (DDPG) and Soft Actor-Critic (SAC) - when performing tasks of mapless navigation for mobile robots. We aim to contribute by showing how the neural network architecture influences the learning itself, presenting quantitative results based on the time and distance of navigation of aerial mobile robots for each approach. Overall, our analysis of six distinct architectures highlights that the stochastic approach (SAC) better suits with deeper architectures, while the opposite happens with the deterministic approach (DDPG).
\end{abstract}


%% file: sections/0_1_supplementary_material.tex
\section*{Supplementary Material}\label{supplementary_material}


%% file: sections/1_introduction.tex
\section{Introduction}
\label{introduction}


Many problems in robotics can be expressed as Reinforcement Learning (RL) problems. RL techniques allow a robot to learn progressively to excel in a distinct task, such as motion-based tasks. Through trial-and-error interactions with an environment, an agent gets feedback in terms of a scalar objective function that guides it step-by-step towards the learning \cite{kober2013reinforcement}. This can be approached by many policies of learning, which can framed into two groups: deterministic and stochastic. 

More recently, RL techniques have been further improved by using deep neural networks. In this case, the agent of Deep Reinforcement Learning (Deep-RL) becomes a neural network that escalates its ability to learn complex behaviors, such as the behavior needed to perform navigation tasks in complex environments. The techniques based on Deep-RL have been used extensively to improve navigation-related tasks for a range of mobile vehicles, including terrestrial mobile robots \cite{ota2020efficient, jesus2021soft}, aerial robots \cite{tong2021uav,grando2022double} and underwater robots \cite{carlucho2018}. These approaches diverge on the choice of an ANN, ranging from Multi-Layer Perceptron (MLP) network structures to Convolutional Neural Networks (CNN). Most of them have achieved interesting results not only in performing mapless navigation-related tasks but also in obstacle avoidance and even media transitioning for hybrid vehicles \cite{bedin2021deep, de2022depth}. However, the choice of the learning method can be affected not only by the selection of the Deep-RL technique but also by the agent's ANN structure.


In this work, we further explore the use of distinct Deep ANN architectures for two state-of-art Deep-RL algorithms: Deep Deterministic Policy Gradient (DDPG) \cite{lillicrap2015continuous} and Soft Actor-Critic (SAC) \cite{haarnoja2018soft}; a deterministic one and a stochastic one. We perform a study showing how the deep network structure and the depth of the network impact the agent's learning. We perform an evaluation using an Unmanned Aerial Vehicle (ANN) in tasks related to goal-oriented mapless navigation using low-dimensional data. We perform a two-fold evaluation, taking into account navigation time and distance navigated to enrich the results.

This work contains the following main contributions:

\begin{itemize}

\item We present comparative analyses of how the agent's deep neural network affects deterministic and stochastic algorithms' performance for mapless navigation-related tasks of mobile robots.

\item We show that low dimensional sensing is better suited to use in Deep-RL for continuous control tasks in general   

\item We also show that the depth increases the performance of deterministic approaches in general, while the opposite tends to happen with stochastic approaches.  

\item We provide a framework with a simulation and environment and two approaches based on state-of-the-art actor-critic Deep-RL algorithms with a range of structures that can be successfully adapted to perform mapless navigation of mobile robots, using only range data readings and the vehicles' relative localization data. 

\end{itemize}

This paper is organized as follows: the related works section (Sec. \ref{related_works}) is presented in the sequence. We show our methodology in Sec. \ref{methodology} and the results are presented in Sec. \ref{results}; both are complemented in Sec. \ref{discussion}. For last, we highlight our contributions and present future works in Sec. \ref{conclusion}.

%% file: sections/2_related_works.tex
\section{Related Work}
\label{related_works}



A couple of Deep-RL works in robotics have already been presented, discussing how efficiently these methods can be used in problems related to motion control with low dimensional sensing information \cite{tai2017virtual, bedin2021deep}. For a terrestrial mobile robot, Tai \emph{et al.} \cite{tai2017virtual} used ten samples of range findings and the relative distance of the vehicle to a target to perform navigation through obstacles. The DDPG algorithm used learned effectively to navigate to a target. Recently, deep-RL methods have also been successfully used in robotics by 
De Jesus \emph{et al.}. Also, \cite{jesus2019deep,jesus2021soft} and others accomplished mapless navigation-related tasks for terrestrial mobile robots using simple information. 

Singh and Thongam \cite{singh2018mobile} show that a Multi-Layer Perceptron (MLP) can be used for mapless navigation of terrestrial mobile robots in dynamic environments. Their method used MLP and Recurrent Neural Networks to decide the robot's speed for each motion. They concluded that the approach is efficient in guiding the robot to a target position. 

For aerial mobile robots, the use of Deep-RL is still limited. Rodriguez \emph{et al.} \cite{rodriguez2018deep} used a DDPG-based approach to teach an agent to land on a moving platform. Their approach used information from images, but it was fed with simplified information to the agent. It used Deep-RL in simulation with the RotorS framework \cite{furrer2016rotors} and with the Gazebo simulator. Grando \emph{et al.} \cite{grando2020deep} presented a DDPG and a SAC approach on Gazebo for 2D UAV navigation. Recently, double critic-based Deep-RL has also been used for UAVs \cite{grando2022double}. All of them use information from ranging sensors in a simple state information model for the agent.

Two works have recently tackled the navigation problem with the medium transition of hybrid mobile robots \cite{de2022depth}, \cite{bedin2021deep}. Grando \emph{et al.} \cite{bedin2021deep} presented Deep-RL approaches with a MLP architecture. It was developed using distance sensing information for aerial and underwater navigation. De Jesus \emph{et al.} \cite{de2022depth} tackled the problem of motion for this kind of vehicle using image information, but it was used with contrasting learning that takes into account a decoder to simplify the image information to feed the agent.

Based on these works that used simple state information, we present a comparative analysis of how the agent's deep neural network affects the performance of Deep-RL for continuous motion tasks in mobile robots. We aim to provide the best architecture for each kind of algorithm. The environment used for the testing is the Gazebo simulator with a described real-world aerial mobile robot.

%% file: sections/3_methodology.tex
\section{Methodology}
\label{methodology}

In this section, we discuss the Deep-RL approaches used in this work and the mobile aerial robot used. We detail the structure of all networks used to perform the comparison for both deterministic and stochastic agents. 

\subsection{Deep Deterministic Policy Gradient}

Deep Deterministic Policy Gradient (DDPG)~\cite{lillicrap2015continuous} has two main deep neural networks: an actor network that provides the real value of a chosen action and a second deep neural network to learn a target function that gives stability to the learning process~\cite{lillicrap2015continuous}. The observation of the current state is the input of the actor-network. The actor network provides a continuous action space value chosen by the policy. At the same time, the critic network uses the current state and the agent's action to provide the Q value for the agent. 

This method provides good performance for continuous control but has a challenging problem related to exploration. Since it is deterministic, DDPG needs some exploration policy $\mu'$ to avoid learning stagnation. This can be solved by adding a noise process $\mathcal{N}$ to the actor policy. This noise adding process can be defined as: 

\begin{equation}
\mu' = \mu(s_t) + \mathcal{N}
\end{equation}
where $\mathcal{N}$ is a noise chosen.
The Ornstein-Uhlenbeck process \cite{uhlenbeck1930theory} is typically used and provides good exploration.

\subsection{Soft Actor Critic (SAC)}

We also developed a stochastic approach for comparison. It was based on the Soft Actor-Critic algorithm~\cite{haarnoja2018soft}. It also consists of an actor-critic system that combines off-policy updates but with a stochastic actor-critic method to learn continuous action space policies. It does so by using neural networks as an approximation function to learn a policy. However, in the SAC algorithm, the current stochastic policy is used instead of the noise used in DDPG. By acting without noise, it tends to provide better stability and performance. The learning speed also tends to be higher since the algorithm encourages the agent to explore new states. It uses the Bellman equation with neural networks as a function approximation to maximize the entropy.

\begin{figure*}[!h]
  \subfloat[MLP 2.\label{fig:mlp_2layers_structure}]{
	\begin{minipage}[c][0.8\width]{
	   0.3\textwidth}
	   \centering
	   \includegraphics[width=1.0\textwidth]{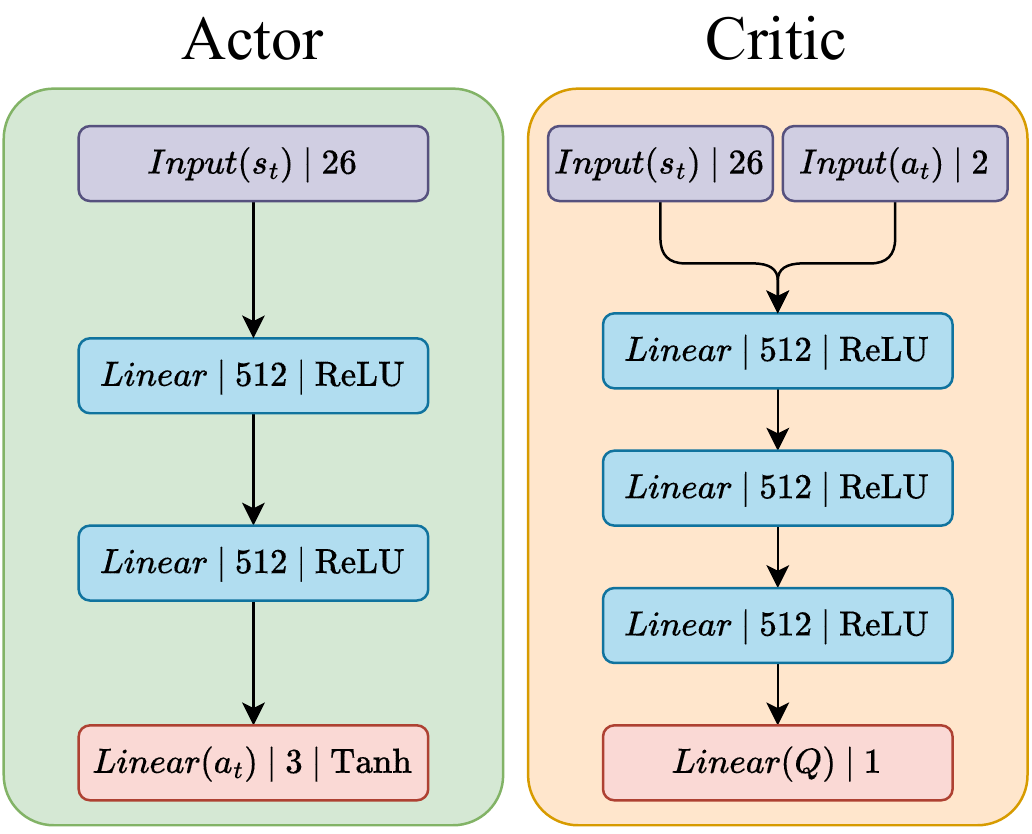}
	\end{minipage}}
 \hfill 	
  \subfloat[MLP 3.\label{fig:mlp_3layers_structure}]{
	\begin{minipage}[c][0.8\width]{
	   0.3\textwidth}
	   \centering
	   \includegraphics[width=1.0\textwidth]{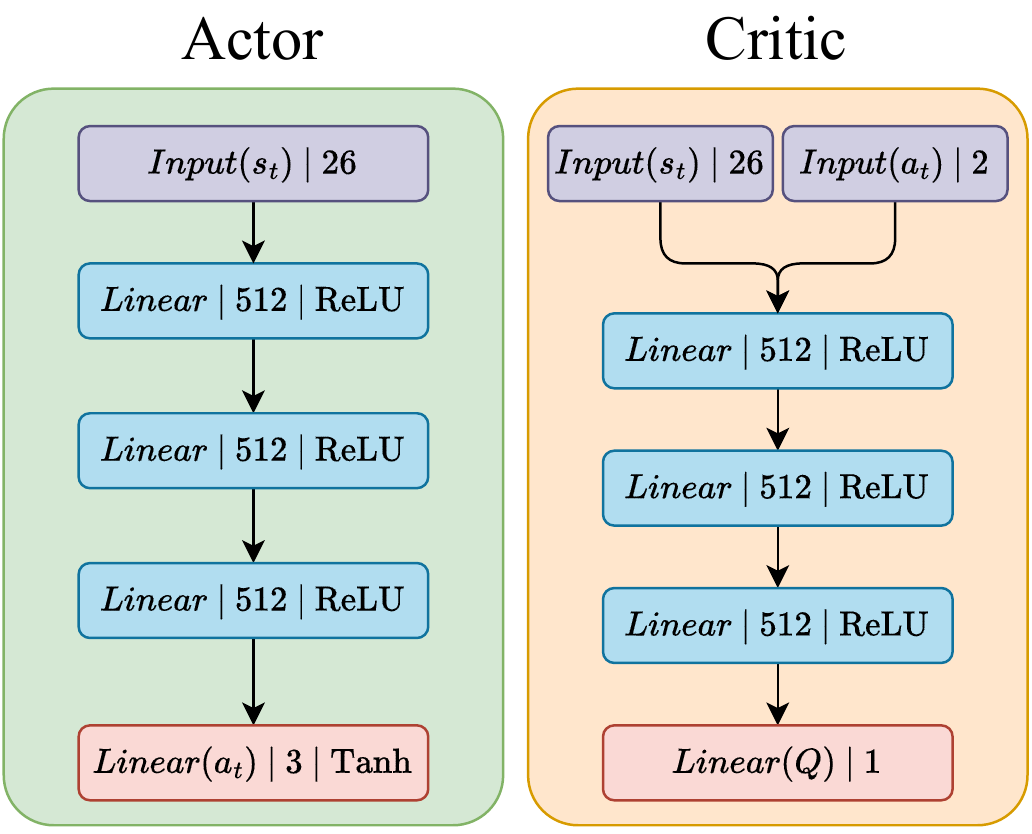}
	\end{minipage}}
 \hfill 
 \subfloat[MLP 4.\label{fig:mlp_4layers_structure}]{
	\begin{minipage}[c][0.8\width]{
	   0.3\textwidth}
	   \centering
	   \includegraphics[width=1.0\textwidth]{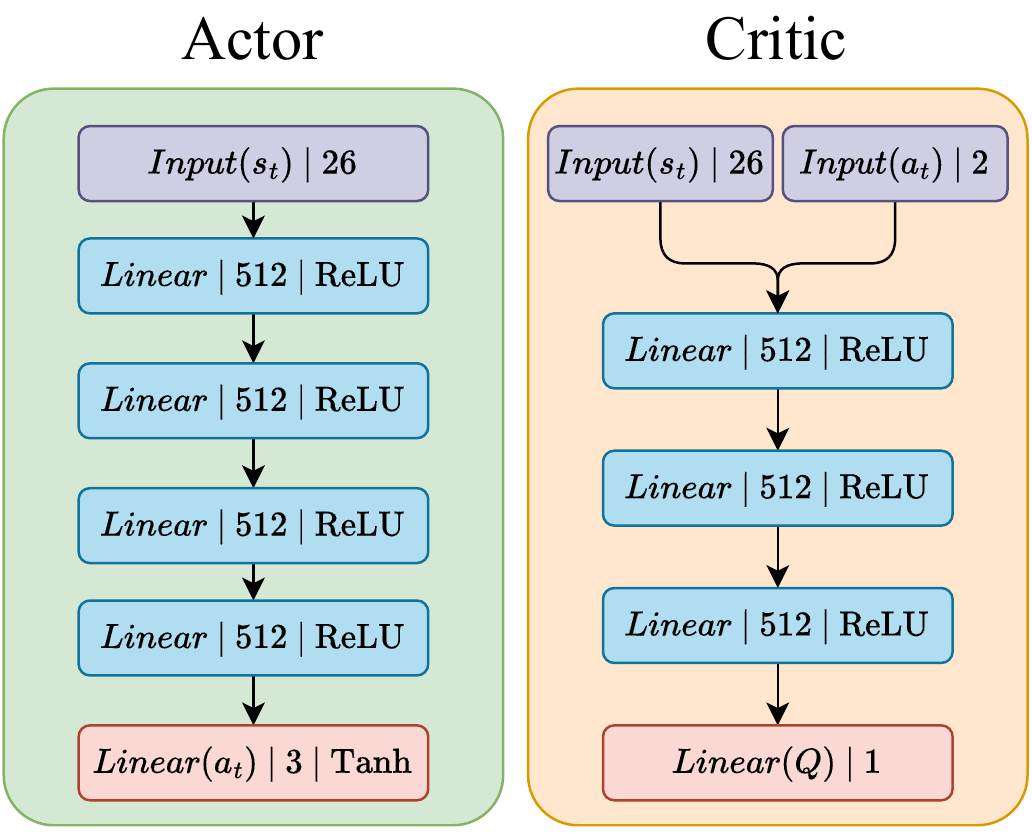}
	\end{minipage}}
 \hfill 
 \subfloat[MLP 5.\label{fig:mlp_5layers_structure}]{
	\begin{minipage}[c][0.8\width]{
	   0.3\textwidth}
	   \centering
	   \includegraphics[width=1.0\textwidth]{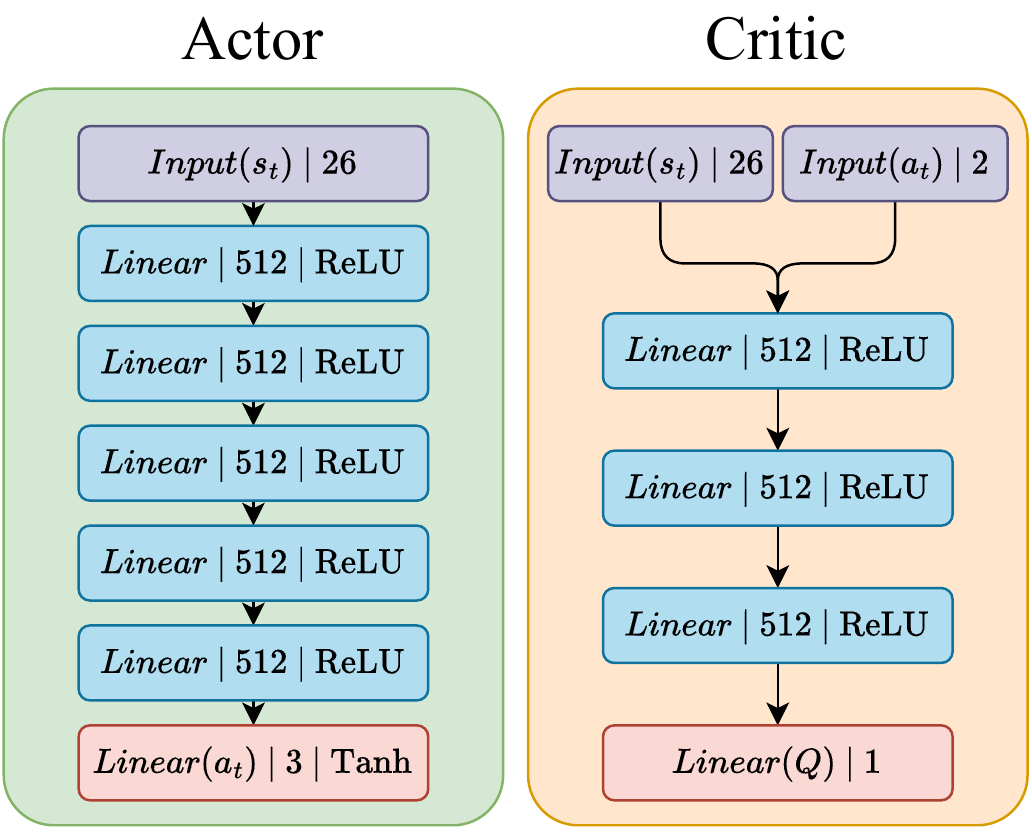}
	\end{minipage}}
 \hfill 
 \subfloat[LSTM.\label{fig:lstm_structure}]{
	\begin{minipage}[c][0.8\width]{
	   0.3\textwidth}
	   \centering
	   \includegraphics[width=1.0\textwidth]{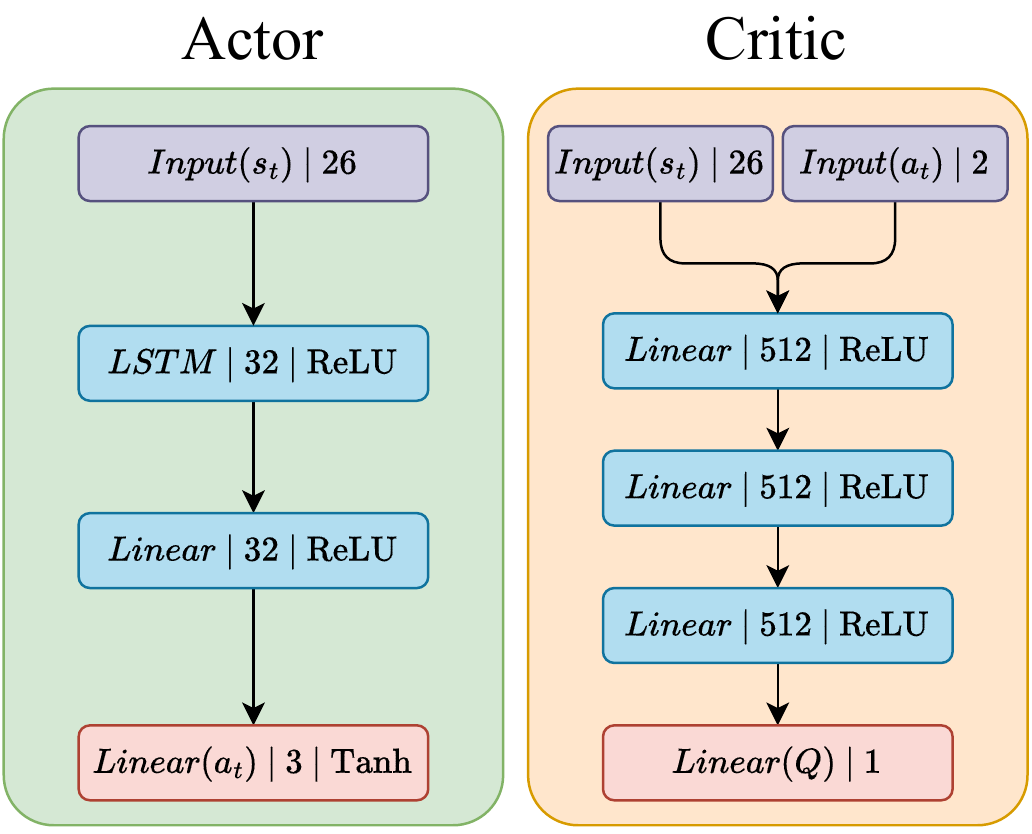}
	\end{minipage}}
 \hfill
 \subfloat[CONVNET.\label{fig:conv_structure}]{
	\begin{minipage}[c][0.8\width]{
	   0.3\textwidth}
	   \centering
	   \includegraphics[width=1.0\textwidth]{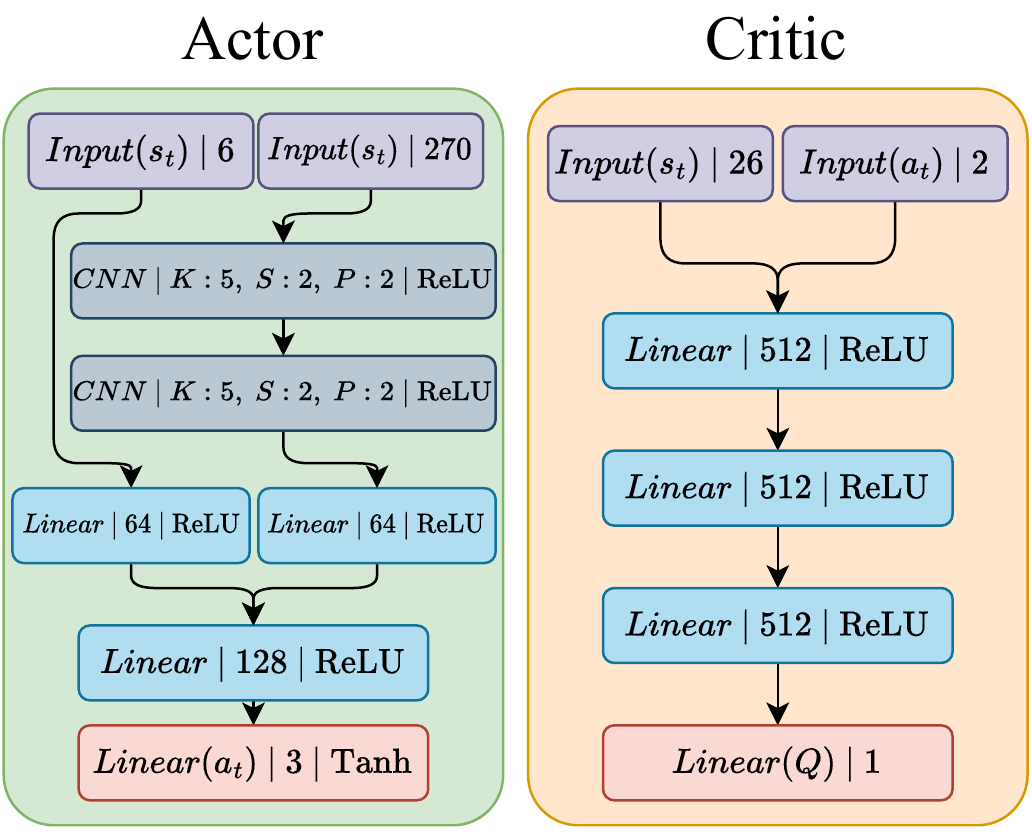}
	\end{minipage}}
 \hfill
\caption{Networks structures used in our comparison study.}
\label{fig:networks_ablation}
\end{figure*}

\subsection{Simulated Environments}

The algorithms were implemented in simulation together with ROS and the Gazebo simulator. We used the aerial mobile vehicle presented at ~\cite{grando2022double}. This vehicle was described using the framework RotorS \cite{furrer2016rotors} and was based on the real vehicle presented at \cite{grando2020visual}. Its low dimensional sensing was given by a simulated LIDAR. The described LIDAR is based on the UST 10LX model. It provides a 10 meters distance sensing with 270\degree~range and 0.25\degree~of resolution. 

We developed two environments with dimensions of 10$\times$10$\times$6 meters. The first one is a simple environment that has no obstacles. It only has walls that limit the scenario. Our idea for using this scenario was to guarantee that all the versions of the ANNs used were able to learn to perform the task of navigating to a target point to have a fair comparison. The second one has a few obstacles added to it to make it more difficult to learn and test the ability of the agent in avoiding obstacles.

\subsection{Reward Function} 

A simple binary rewarding function was used: a positive reward is given if the agent reaches the goal, or a negative reward is given if the robot collides with the walls or does not reach the goal in less than a 500 steps limit. The function can be described as follows:

\vspace{-5mm}
\begin{equation}
r(s_t, a_t)= 
\begin{cases}
    r_{arrive}           & \text{if } d_t < c_d\\
    r_{collide}          & \text{if } min_x < c_o\ ||\ ep = 500\\
\end{cases}
\end{equation}

The reward $r_{arrive}$ is  100, while the negative reward $r_{collide}$ is -10. Both $c_d$ and $c_o$ distances were set to $0.5$ meters.

\subsection{Networks Structure} \label{secapproach}

The input for the networks has a total of 26 values, 20 samples for the distance sensors, the three previous actions and three values related to the target goal, which are the vehicle's relative position to the target and the relative angle to the target in the x-y plane and z-distance plane. The only exception was used in the CNN architecture, where 270 samples from the sensor were used instead of 20. The network outputs are the linear velocity and the variation of the vehicle's yaw ($\Delta$ yaw) that will be sent to the robot. The actions are normalized between $0$ and $0.25$ $m/s$ for the linear velocity and from $-0.25$ to $0.25$ $rad$ for the $\Delta$ yaw. All network structures were developed inspired by related works that deal with low-dimensional sensing data as inputs.

We used six distinct ANN architectures, four of them based on the MLP fully connected architecture, one based on the LSTM architecture and one based on a CNN architecture. For all actor-critic network structures, we fixed the critic network with a standard of three hidden layers with 512 neurons each, while the actor-network varies, as shown in Figure \ref{fig:networks_ablation}. The main idea is to evaluate the impact of the network structure on the agent's ability to provide the actions, not to evaluate if the agent is capable of learning, which is the main goal of the critic network. For the last, we used ReLU activation in the hidden layers and Tanh activation in the output layer.

%% file: sections/4_experimental_results.tex
\section{Experimental Results}
\label{results} 

For training, we generated target goals in a random manner, towards which the agent should navigate. Five hundred steps was the limit defined for each episode, which could end first if the agent collided with an obstacle or with the scenario border. A new goal in the same episode was generated if the agent reached the goal before finishing the 500 steps limit. In this case, the total amount of reward could exceed the maximum value of 100.A learning rate of $10^{-3}$ was used, with a minibatch of 256 samples and the Adam optimizer for all approaches. In the first scenario, we limited the number of episodes to be trained to 1000, while the agent was trained for 1500 episodes in the second. These respective limits for the episode number are used based on the stagnation of the maximum average reward received.

\subsection{Results} 

In this section, the results obtained during our evaluation are shown. For each scenario and model, an extensive amount of statistics are collected. The evaluation was done in a two-fold manner, with goal-oriented navigation and a waypoint navigation manner. These two-fold tasks were performed for 100 trials and the total of successful trials are recorded. Also, the average navigation time with its standard deviation is recorded. 


\begin{table}[!h]
\centering
\caption{Statistics of goal-oriented navigation 2D}
\label{table:nav_air_2D_ablation_ddpg}
\begin{tabular}{c c c c} 
\toprule
Env & Algorithm & Rate & Average Time (s) \\
\midrule
1 & DDPG 2  & 100\% & $18.54 \pm 1.21$ \\
1 & DDPG 3  & 100\% & $13.48 \pm 0.46$ \\
1 & DDPG 4  & 100\%& $13.442 \pm 0.96$ \\
1 & DDPG 5  & 0\% & $102.812 \pm 10.849 $ \\
1 & DDPG LSTM  & 95\% & $58.657 \pm 65.295$ \\
1 & DDPG CONV  & 84\% & $48.329 \pm 48.522$ \\
2 & DDPG 2   & 65\% & $30.711 \pm 11.343$ \\
2 & DDPG 3  & 35\% & $28.13 \pm 11.06$ \\
2 & DDPG 4  & 39\% & $29.9 \pm 14.62$ \\
2 & DDPG 5  & 0\% & $92.547 \pm 17.86$ \\
2 & DDPG LSTM  & 21\% & $44.06 \pm 26.49$ \\
2 & DDPG CONV  & 0\% & $98.77 \pm 82.11$ \\
1 & SAC 2  & 100\% & $18.38 \pm 1.05$ \\
1 & SAC 3  & 93\% & $32.2 \pm 14.52$ \\
1 & SAC 4  & 100\% & $16.16 \pm 1.63$ \\
1 & SAC 5  & 97\% & $24.87 \pm 14.896 $ \\
1 & SAC LSTM  & 100\% & $19.22 \pm 4.37$ \\
1 & SAC CONV  & 100\% & $15.233 \pm 1.023$ \\
2 & SAC 2   & 13\% & $39.99 \pm 26.32$ \\
2 & SAC 3  & 74\% & $35.617 \pm 12.79$ \\
2 & SAC 4  & 60\% & $33.978 \pm 11.84$ \\
2 & SAC 5  & 32\% & $25.011 \pm 12.29$ \\
2 & SAC LSTM  & 35\% & $39.87 \pm 20.71$ \\
2 & SAC CONV  & 0\% & $34.01 \pm 14.67$ \\
\bottomrule
\end{tabular}
\end{table}

\begin{table}[!h]
\centering
\caption{Statistics of waypoint goal-oriented navigation 2D}
\label{table:multinav_air_2D_ablation_ddpg}
\begin{tabular}{c c c c c } 
\toprule
Env & Algorithm & Rate & Average Time (s) & $\%$ Distance \\
\midrule
1 & DDPG 2  & 100\% & $97.22\pm 1.79$ & 100\% \\
1 & DDPG 3  & 100\% & $99.37 \pm 3.89$ & 100\% \\
1 & DDPG 4  & 95\% & $135.84 \pm 29.146$ & 97.376\% \\
1 & DDPG 5  & 0\% & $203.168 \pm 1.849  $ & 0\% \\
1 & DDPG LSTM  & 84\% & $248.04 \pm 167.75$ & 87.75\%\\
1 & DDPG CONV  & 39\% & $382.637 \pm 211.715$ & 67.89\%\\
2 & DDPG 2   & 1\% & $118.278 \pm 75.77$ & 46.734\%\\
2 & DDPG 3  & 7\% & $55.43 \pm 44.319$ & 20.816\% \\
2 & DDPG 4  & 0\% & $45.09 \pm 35.45$ & 10\%\\
2 & DDPG 5  & 0\% & $97.54 \pm 16.79$ & 0\%\\
2 & DDPG LSTM  & 0\% & $75.06 \pm 53.6$ & 7.142\%\\
2 & DDPG CONV  & 0\% & $127.50 \pm 74.74$ & 0.61\%\\
1 & SAC 2  & 2\% & $131.07 \pm 88.29$ & 32.21\% \\
1 & SAC 3  & 54\% & $164.18 \pm 43.89$ & 73.76\% \\
1 & SAC 4  & 68\% & $155.54 \pm 40.36$ & 83.09\% \\
1 & SAC 5  & 48\% & $152.418 \pm 50.384$ & 74.63\% \\
1 & SAC LSTM  & 100\% & $118.55 \pm 11.16$ & 100\%\\
1 & SAC CONV  & 0\% & $209.74 \pm 2.051$ & 50.43\%\\
2 & SAC 2   & 0\% & $53.379 \pm 39.343$ & 11.224\%\\
2 & SAC 3  & 0\% & $36.611 \pm 16.65$ & 12.65\% \\
2 & SAC 4  & 0\% & $54.508 \pm 42.59$ & 14.285\%\\
2 & SAC 5  & 9\% & $73.56 \pm 59.489$ & 28.775\%\\
2 & SAC LSTM  & 1\% & $62.92 \pm 46.63$ & 12.85\%\\
2 & SAC CONV  & 0\% & $35.6 \pm 15.30$ & 0.204\%\\
\bottomrule
\end{tabular}
\end{table}

\begin{table}[!h]
\centering
\caption{Statistics of goal-oriented navigation 3D}
\label{table:nav_air_3D_ablation_ddpg}
\begin{tabular}{c c c c} 
\toprule
Env & Algorithm & Rate & Average Time (s) \\
\midrule
1 & DDPG 2  & 100\% & $13.54 \pm 1.00$ \\
1 & DDPG 3  & 100\% & $13.697 \pm 0.85$ \\
1 & DDPG 4  & 100\% & $13.735 \pm- 0.941$ \\
1 & DDPG 5  & 0\% & $20.21 \pm 13.93 $ \\
1 & DDPG LSTM  & 100\% & $13.81 \pm 1.22$ \\
1 & DDPG CONV  & 0\% & $26.66 \pm 17.83$ \\
2 & DDPG 2   & 8\% & $23.187 \pm 10.30$ \\
2 & DDPG 3  & 17\% & $49.276 \pm 38.32$ \\
2 & DDPG 4  & 24\% & $29.76 \pm 15.85$ \\
2 & DDPG 5  & 0\% & $17.93 \pm 5.727$ \\
2 & DDPG LSTM  & 61\% & $31.26 \pm 16.43$ \\
2 & DDPG CONV  & 0\% & $18.56 \pm 10.57$ \\
1 & SAC 2  & 100\% & $25.89 \pm 10.39$ \\
1 & SAC 3  & 100\% & $27.038 \pm 18.673$ \\
1 & SAC 4  & 91\% & $26.008 \pm 15.20$ \\
1 & SAC 5  & 100\% & $16.706 \pm 2.674$ \\
1 & SAC LSTM  & 100\% & $56.48 \pm 25.71$ \\
1 & SAC CONV  & 0\% & $68.75 \pm 51.81$ \\
2 & SAC 2   & 5\% & $28.218 \pm 16.553$ \\
2 & SAC 3  & 19\% & $39.43 \pm 22.21$ \\
2 & SAC 4  & 29\% & $50.373 \pm 40.765$ \\
2 & SAC 5  & 8\% & $51.58 \pm 23.25$ \\
2 & SAC LSTM  & 5\% & $37.44 \pm 24.72$ \\
2 & SAC CONV  & 0\% & $32.78 \pm 17.62$ \\
\bottomrule
\end{tabular}
\end{table}

\begin{table}[!h]
\centering
\caption{Statistics of waypoint goal-oriented navigation 3D}
\label{table:multinav_air_3D_ablation_ddpg}
\begin{tabular}{c c c c c } 
\toprule
Env & Algorithm & Rate & Average Time (s) & $\%$ Distance \\
\midrule
1 & DDPG 2  & 100\% & $72.90 \pm 2.21$ & 100\% \\
1 & DDPG 3  & 77\% & $76.99 \pm 29.23$ & 86.44\% \\
1 & DDPG 4  & 99\% & $90.120 \pm 19.90$ & 98.396\% \\
1 & DDPG 5  & 0\% & $19.529 \pm 20.32$ & 1.02\% \\
1 & DDPG LSTM  & 99\% & $90.44 \pm 6.02$ & 98.97\%\\
1 & DDPG CONV  & 0\% & $20.53 \pm 11.1790$ & 0.62\%\\
2 & DDPG 2   & 0\% & $29.194 \pm 23.434$ & 2.24\%\\
2 & DDPG 3  & 0\% & $39.8 \pm 28.85$ & 2.24\% \\
2 & DDPG 4  & 0\% & $38.41 \pm 31.38$ & 4.897\%\\
2 & DDPG 5  & 0\% & $17.95 \pm 6.486$ & 0\%\\
2 & DDPG LSTM  & 0\% & $34.9 \pm 25.37$ & 8.36\%\\
2 & DDPG CONV  & 0\% & $21.01 \pm 18.51$ & 0\%\\
1 & SAC 2  & 0\% & $114.01 \pm 67.66$ & 37.463\% \\
1 & SAC 3  & 10\% & $135.220 \pm 61.154$ & 47.084\% \\
1 & SAC 4  & 52\% & $140.94 \pm 50.30$ & 68.22\% \\
1 & SAC 5  & 26\% & $173.124 \pm 48.973$ & 55.685\% \\
1 & SAC LSTM  & 94\% & $155.82 \pm 60.69$ & 95.34\%\\
1 & SAC CON  & 0\% & $121.21 \pm 73.98$ & 1.603\%\\
2 & SAC 2   & 0\% & $25.054 \pm 15.52$ & 0.204\%\\
2 & SAC 3  & 0\% & $42.44 \pm 31.20$ & 3.469 \%\\
2 & SAC 4  & 3\% & $61.382 \pm 55.434$ & 16.938\%\\
2 & SAC 5  & 1\% & $57.499 \pm 28.62$ & 3.06\%\\
2 & SAC LSTM  & 0\% & $34.9 \pm 25.37$ & 1.224\%\\
2 & SAC CONV  & 0\% & $29.46 \pm 12.94$ & 0\%\\
\bottomrule
\end{tabular}
\end{table}

%% file: sections/5_discussion.tex
\section{Discussion}
\label{discussion}

\begin{figure*}[!h]
  \subfloat[Reward 2D Navigation Scenario 1.\label{fig:reward_all_2d_1}]{
	\begin{minipage}[c][0.7\width]{0.49\textwidth}
	   \centering
	   \includegraphics[width=\textwidth]{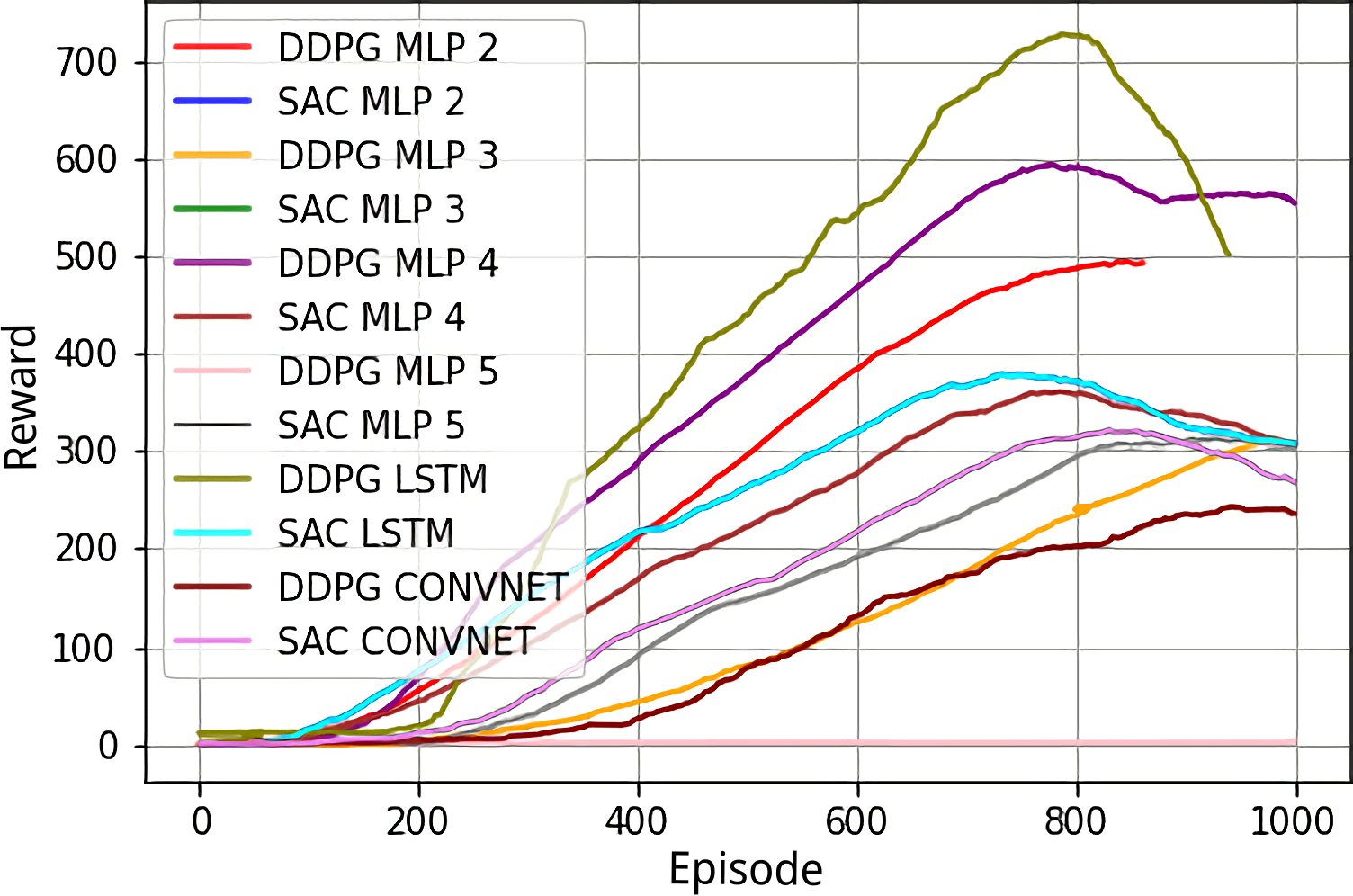}
	\end{minipage}}
 \hfill 	
  \subfloat[Reward 2D Navigation Scenario 2.\label{fig:reward_all_2d_2}]{
	\begin{minipage}[c][0.7\width]{0.49\textwidth}
	   \centering
	   \includegraphics[width=\textwidth]{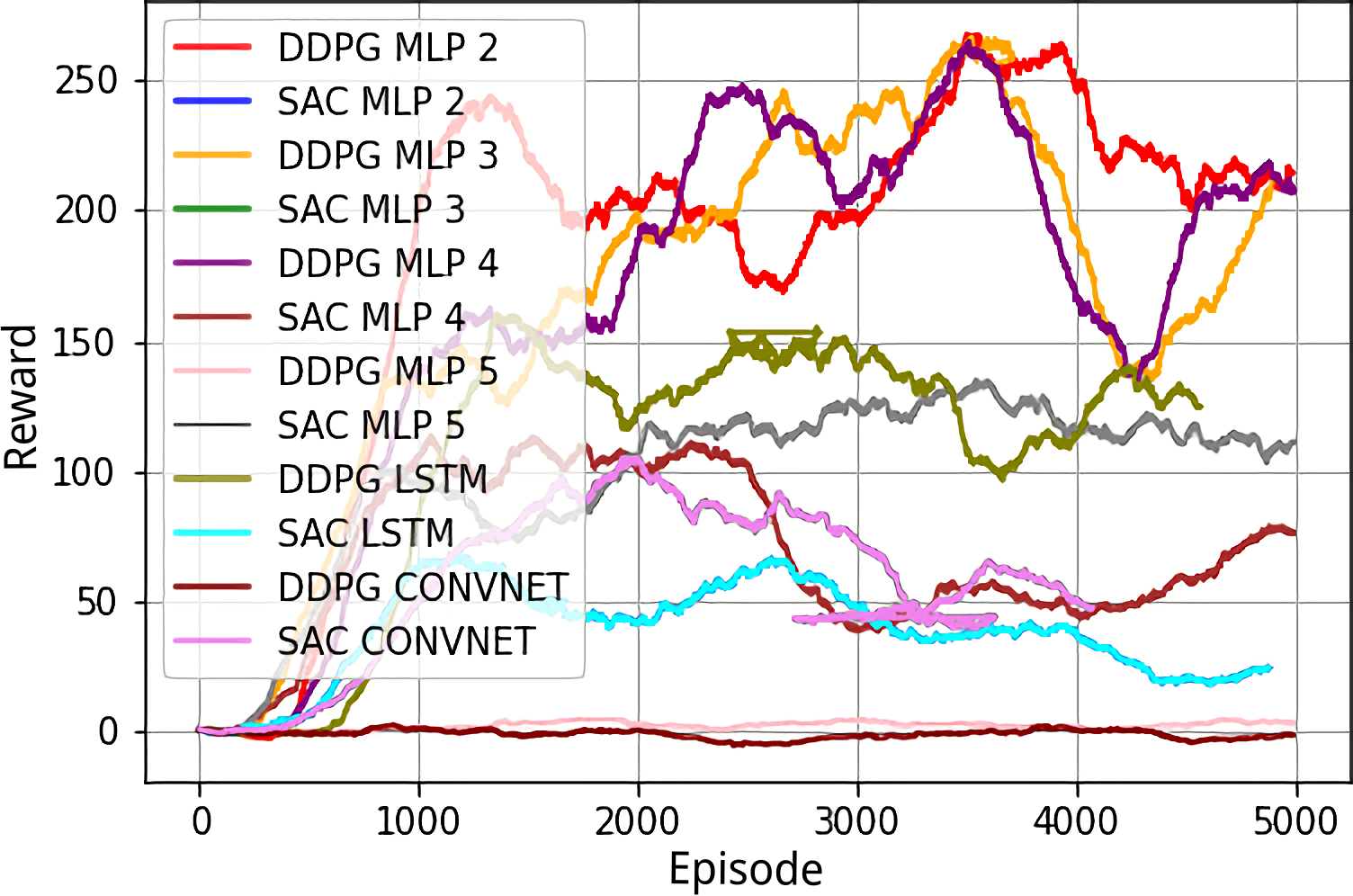}
	\end{minipage}}
 \hfill 
 \subfloat[Reward 3D Navigation Scenario 1.\label{fig:reward_all_3d_1}]{
	\begin{minipage}[c][0.7\width]{0.49\textwidth}
	   \centering
	   \includegraphics[width=\textwidth]{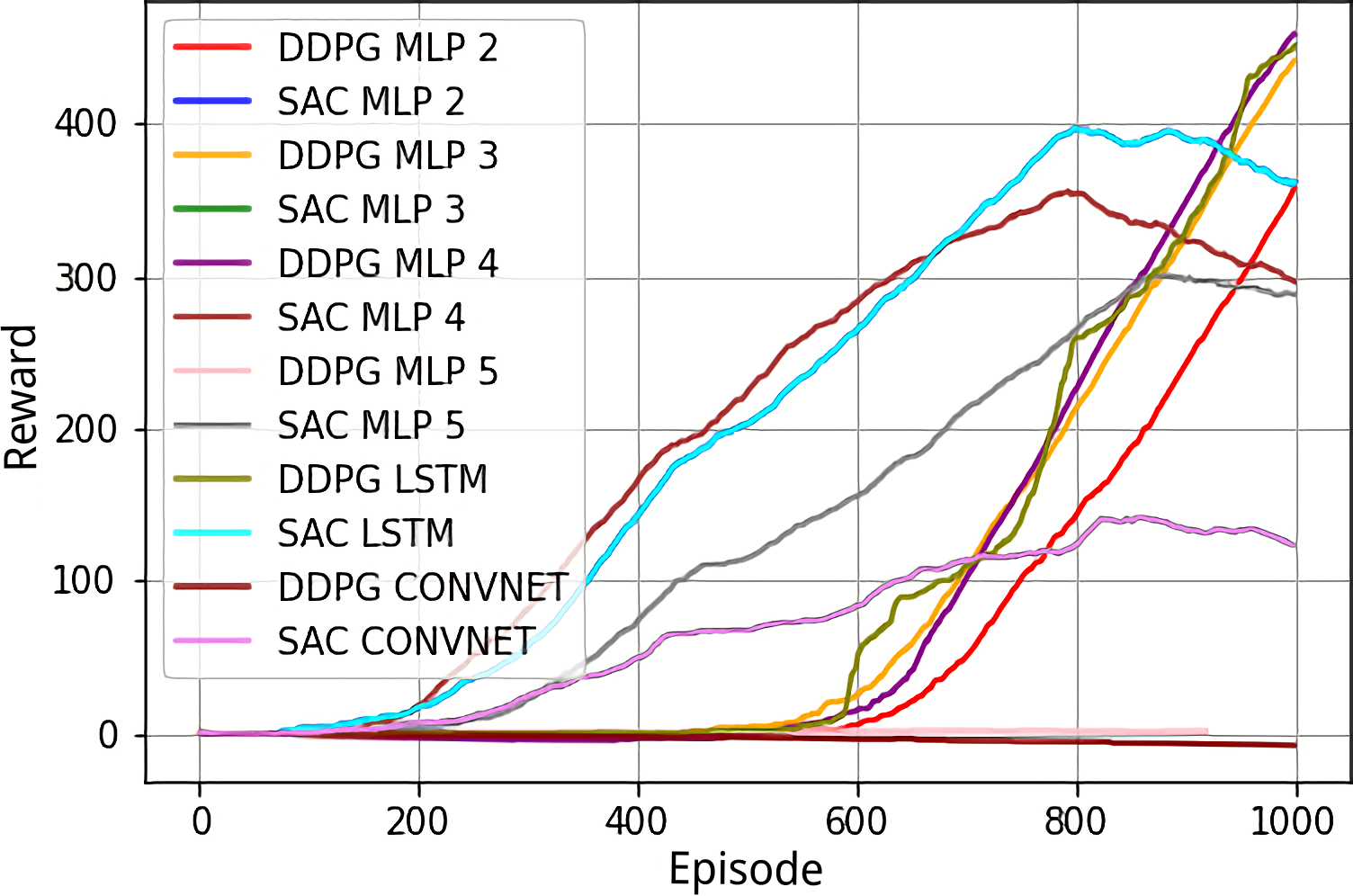}
	\end{minipage}}
 \hfill 	
  \subfloat[Reward 2D Navigation Scenario 2.\label{fig:reward_all_3d_2}]{
	\begin{minipage}[c][0.7\width]{0.49\textwidth}
	   \centering
	   \includegraphics[width=\textwidth]{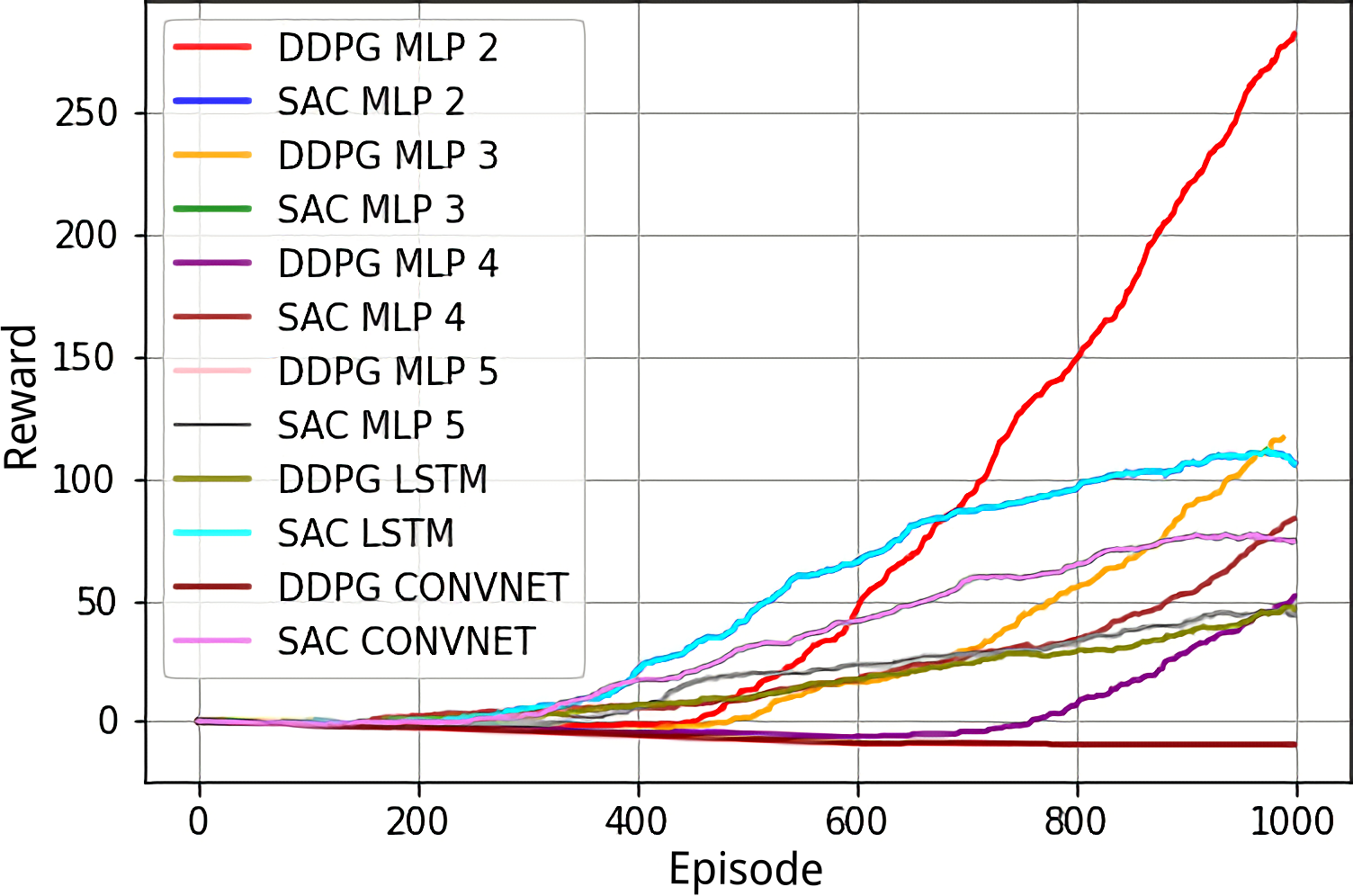}
	\end{minipage}}
 \hfill
\caption{Comparative Reward.}
\label{fig:rewards_all}
\vspace{-5mm}
\end{figure*}

In general, the extensive validation of the various models created and tested shows that both agents are flexible regarding the type of ANNs used. It can be concluded that the DDPG-based approach performs better in an unhindered scenario, while the opposite occurs with the SAC-based approaches. Figure \ref{fig:rewards_all} shows the final reward of each model in each context and scenario.

It can be observed that the larger and more complex the network, the greater the average reinforcement tends to be, as, for example, for the models with LSTM and CNN. It is important to note that this is due to the fact that a greater number of navigations are performed in each episode and not that the models are better or worse. All models with an average reinforcement greater than 100 can be considered functional. The higher number of navigations is due to the slower step with more complex networks, allowing the task to be completed and optimized with a smaller number of actions. This further reinforces the importance of focusing on a simple reward system like the one proposed in this work.

\begin{figure}
\centering
\includegraphics[width=8cm]{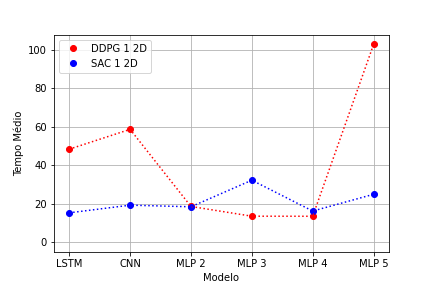}
\caption{Comparison of average time for 2D navigation in the first scenario (more stable).} \label{fig:time_comparision}
\vspace{-5mm}
\end{figure}

In Figure \ref{fig:time_comparision} it is possible to observe the comparison of the average time to perform the first one in the 2D context, the context where the approaches presented average results close to the maximum possible for all structures. It is interesting to observe in Figure \ref{fig:time_comparision} the characteristics of each approach in more detail. It is possible to observe that the SAC approach has a similar average time between the structures, while the DDPG-based approach varies with greater intensity. This is due to the greater generalization capability that the stochastic biased method, such as SAC provides, while DDPG can be very good for specific structures. In general, with respect to time, it is possible to conclude that the approach based on SAC tends to be, on average, a little longer and more predictable, while the opposite occurs with approaches based on DDPG.

\begin{figure}
\centering
\includegraphics[width=8cm]{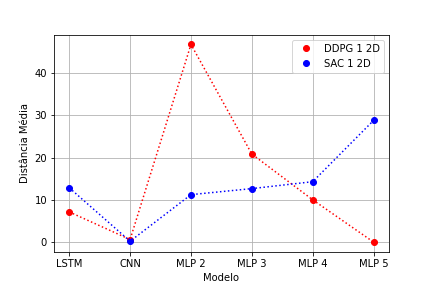}
\caption{Comparison of average time for 2D waypoint navigation in the first scenario (more stable).} \label{fig:distance_comparision}
\vspace{-5mm}
\end{figure}

In Figure \ref{fig:distance_comparision} it is also possible to observe the average distance for the second task in the second scenario, also in the 2D context, as it is more generally stable between the structures. From this illustration, it is interesting to observe in more detail the characteristics of each approach. It can be seen how the DDPG-based approach performs better with two layers and how the performance drops with increasing network complexity. Meanwhile, the SAC approach presents better results with more complex network structures, increasing performance as the number of layers increases, for example. This is due to the ability to generalize and create greater gradients than the method based on stochastic bias SAC has. In general, it can be concluded that the larger the network, the better the performance of agents based on SAC tends to be, while the opposite occurs with DDPG.

The limit for complexity, however, appears to be close to the proposed convolutional model. As can be seen in Figure \ref{fig:distance_comparision} and also in the results for the 3D context, both approaches with CNN failed to learn to perform the tasks. The solution to this can be contrastive networks \cite{de2022depth}. The use of contrastive networks with Deep-RL can be a way not only to solve this problem with CNNs, but also to optimize the problem of the work as a whole.

%% file: sections/6_conclusion.tex
\section{Conclusions}
\label{conclusion}

In this paper, we presented a comparative analysis of deterministic and stochastic algorithms for low-dimensional sensing-based mapless navigation-related tasks for mobile robots. We discussed how the agent's deep neural network affects performance while executing the tasks. We can conclude that the depth of the neural network increases the inefficiency of deterministic approaches in general, while the opposite tends to happen with stochastic approaches. We can also conclude that low-dimensional sensing is better suited to use in Deep-RL for continuous control tasks in general. Overall, future work related to the effect of the critic's neural network will be conducted as well to evaluate how it impacts the learning of the policy itself.

%% file: sections/7_acknowledgment.tex
\section*{Acknowledgment}


The authors would like to thank the VersusAI team. This work was partly supported by the CAPES, CNPq and PRH-ANP.

\vspace{-2mm}

%% file: sections/8_references.tex
\bibliographystyle{./bibliography/IEEEtran}
\bibliography{./bibliography/IEEEabrv,./bibliography/IEEEexample}